\documentclass{article}

\usepackage{amssymb}
\usepackage{bm}
\usepackage{spconf,amsmath,graphicx,hyperref}
\usepackage{booktabs}
\usepackage{multirow}
\usepackage{adjustbox}
\usepackage{makecell}
\newcommand{\confighead}[1]{\hspace*{1.0em}\textit{#1}}
\newcommand{\configmethod}[1]{\hspace*{2.0em}#1}

\title{Language-Specialized Multi-Teacher On-Policy Distillation for Multilingual LLM-Based ASR}
\name{
Yuan Xie$^{\ast}$, 
Jiaqi Song, 
Xianliang Wang, 
Ming Lei, 
Jie Gao, 
Jie Wu 
\thanks{$^{\ast}$ Corresponding author. E-mail: ryan.xie2@nio.com}
}
\address{
Advanced Intelligent Systems Group, NIO, Beijing, China\\[-0.1em]
}

\begin{document}
\ninept
\maketitle
\begin{abstract}
Modern LLM-based ASR systems have established multilingual capability as a standard feature, leveraging large-scale multilingual corpora and LLMs' cross-lingual knowledge to achieve competitive performance across multilingual benchmarks. 
However, jointly modeling languages with heterogeneous acoustic, phonological, and lexical characteristics inevitably introduces optimization conflicts, undermining language-wise specialization. 
To address this challenge, we propose \textbf{Language-Specialized Multi-Teacher On-Policy Distillation (LS-MOPD)}, which decouples language-specific knowledge acquisition from multilingual capability integration: language-specialized teachers are independently optimized via reinforcement learning (RL), with their expertise then integrated into a generalist multilingual student through language routing and token-level multi-teacher distillation, thereby reducing direct cross-lingual optimization conflicts. We further explore static and dynamic acoustic-prefix configurations to examine how teacher--student prefix consistency influences the efficacy of on-policy distillation. 
Experiments on benchmarks covering Mandarin, Mandarin subdialects, Cantonese, and English demonstrate that LS-MOPD substantially outperforms RL baselines and surpasses the empirical performance envelope defined by the best-performing RL teachers on nearly all benchmarks, revealing its potential to \textbf{generalize beyond all teachers} in multilingual ASR.

\end{abstract}
\begin{keywords}
multilingual ASR, speech large language models, multi-teacher on-policy distillation
\end{keywords}
\section{Introduction}
\label{sec1}

LLM-based ASR has emerged as a mainstream paradigm in speech recognition, capitalizing on the extensive model capacity and rich linguistic priors of pretrained LLMs to achieve impressive performance across dozens of languages and dialects. Frontier systems such as Seed-ASR~\cite{bai2024seed}, Fun-ASR~\cite{an2025fun}, and Qwen3-ASR~\cite{shi2026qwen3} primarily rely on scaling up multilingual corpora, integrating broad language capabilities into a generalist model in a data-driven manner.

Despite advances on multilingual ASR benchmarks, consolidating cross-lingual capabilities within a generalist model remains challenging. Languages and dialects differ systematically in phonological, acoustic, and linguistic properties~\cite{farooq2022investigating}. For instance, Mandarin is characterized by pervasive homophony and thus relies heavily on contextual semantic disambiguation, whereas English exhibits considerable pronunciation and prosodic variability, demanding precise modeling of stress-timed rhythm and connected-speech phenomena such as phonetic reduction and cross-word linking. Jointly modeling such heterogeneous languages inevitably introduces gradient conflicts, causing cross-lingual performance trade-offs~\cite{wang2020negative}. This challenge intensifies in real-time ASR systems, where latency constraints limit model scale, requiring heterogeneous multilingual capabilities to be accommodated within a finite model capacity budget.

Recent advances in multi-teacher on-policy distillation (MOPD) offer a principled remedy to this challenge~\cite{xiao2026mimo,ma2026mopd}. 
MOPD first trains multiple task-specialized teachers and distills their expertise into a student through task-routed supervision on the student's on-policy rollouts~\cite{yang2026nemotron,xu2026deepseek}. In multilingual ASR, task routing can be naturally adapted to language routing, decoupling language-specific optimization from multilingual integration. Heterogeneous language-wise objectives are independently optimized by specialized teachers, then distilled into the student, mitigating direct cross-lingual optimization conflicts. 
However, extending MOPD to LLM-based ASR raises several unexplored issues. Unlike text-only LLMs, LLM-based ASR conditions decoding on continuous acoustic prefixes produced by the encoder--adaptor. The teacher--student prefix consistency depends on teacher training strategy and may directly influence distillation efficacy. Moreover, cross-lingual heterogeneity is generally less pronounced than cross-task heterogeneity, leaving it unclear whether aggregating supervision from multiple language-specialized teachers can yield genuinely complementary gains in multilingual ASR.

\begin{figure*}[ht]
    \centering
    \includegraphics[width=1.0\linewidth]{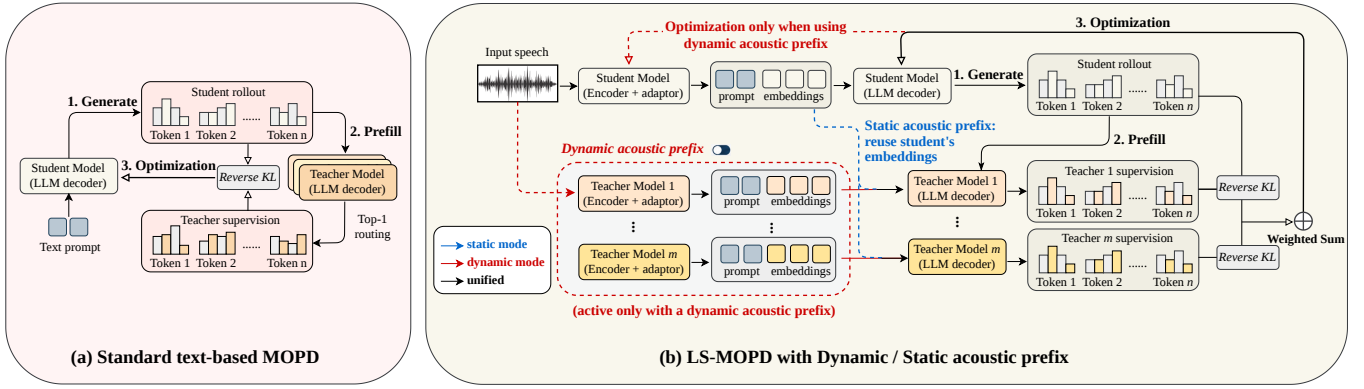}
    \caption{Comparison of (a) standard text-based MOPD and (b) LS-MOPD for ASR with static and dynamic acoustic prefixes. With dynamic prefixes, teachers and the student use independently optimized encoder--adaptors, yielding model-specific acoustic prefixes; with static prefixes, all models share a frozen encoder--adaptor, allowing teachers to reuse the student's acoustic prefixes.}
    \label{fig1}
    \vspace{-3px}
\end{figure*}

To address these issues, we investigate alternative MOPD configurations through gradient alignment and multi-teacher complementarity, and propose \textbf{Language-Specialized Multi-Teacher On-Policy Distillation (LS-MOPD)}, the first MOPD framework for ASR. 
Specifically, we first optimize language-specialized teachers using Decoupled Clip and Dynamic Sampling Policy Optimization (DAPO)~\cite{yu2026dapo} with an ASR-specific reward. During distillation, each student-generated on-policy rollout is routed to the top-$K$ teachers for its input-language category, with their token-level supervision weighted and aggregated to transfer complementary expertise to the generalist student.  
Extensive experiments on Mandarin, Mandarin subdialects, Cantonese, and English benchmarks demonstrate that LS-MOPD consistently outperforms RL baselines and can generalize beyond the performance envelope of all teachers.

\section{Related Work}
\label{sec2}

\noindent\textbf{Multilingual ASR.} 
Extending multilingual capability has become a major focus of modern ASR systems, with efforts following two main directions. The first leverages industrial-scale data expansion: systems such as Seed-ASR~\cite{bai2024seed}, Fun-ASR~\cite{an2025fun}, Qwen3-ASR~\cite{shi2026qwen3}, FireRedASR2S~\cite{xu2026fireredasr2s}, and NIM4-ASR~\cite{xie2026nim4} achieve strong multilingual performance by combining large-scale multilingual speech corpora with LLMs' cross-lingual priors. However, phonological and lexical heterogeneity across languages generally demands greater model capacity, whereas real-time ASR systems must operate under stringent latency and cost constraints. This tension is reflected in deployment practice, where providers release multilingual-optimized variants, such as Seed-ASR (ML)~\cite{bai2024seed} and Fun-ASR-MLT-nano~\cite{an2025fun}, alongside their flagship models. 
The second direction introduces language-specific architectural mechanisms, such as language-aware adaptors, mixture-of-experts routing, and gated query banks~\cite{li2026mosa,lin2026enhancing,gopal2026language}. While these methods can improve multilingual performance at lower cost, they do not directly enhance the model's underlying acoustic or semantic modeling capacity and may incur additional deployment overhead under high-concurrency inference.

\noindent\textbf{MOPD.} 
On-policy distillation (OPD)~\cite{agarwal2024policy} has recently emerged as an effective LLM post-training algorithm, providing token-level supervision on the student's sampled trajectories to mitigate exposure mismatch. This property makes OPD well suited to LLM-based ASR systems, which perform autoregressive generation yet require strict token-level transcription accuracy.  
Multi-teacher OPD~\cite{xiao2026mimo,ma2026mopd} extends this idea by independently optimizing multiple domain-specialized teachers and distilling them into the student model with reverse-KL-style objectives, making it suitable for capability integration. This is particularly appealing for multilingual ASR, as different languages rely on distinct acoustic-phonetic perception and lexical-semantic modeling capabilities. To the best of our knowledge, we are the first to introduce MOPD into the ASR domain.

\section{Methods}
\label{sec-methods}

\subsection{Language-Specialized RL Teacher Training}
\label{sec:domain_specialized_rl}

Prior to conducting MOPD, we construct a suite of teacher models specialized for distinct language categories.  
Starting from a base model $\pi_{\theta_0}$ built upon the encoder--adaptor--LLM architecture, we perform reinforcement learning with verifiable rewards (RLVR) based on DAPO~\cite{yu2026dapo}. For a given utterance 
$\mathbf{x}$ with reference transcription $y$, the rollout policy $\pi_{\theta_{\mathrm{old}}}$ samples $G=8$ hypotheses, $\{\tau_i\}_{i=1}^{G} \sim \pi_{\theta_{\mathrm{old}}}(\cdot\mid\mathbf{x})$.
Following the reward design adopted in~\cite{xie2026nim4}, each hypothesis is scored by an accuracy reward:
\begin{equation}
R_i
=
R_{\mathrm{acc}}(\tau_i,y)
=
\exp\left[
-2\cdot\,
\operatorname{ER}_{\ell(\mathbf{x})}
\left(\tau_i,y
\right)
\right],
\end{equation}
where $\ell(\mathbf{x})$ denotes the language category of the input and $\operatorname{ER}_{\ell(\mathbf{x})}$ denotes the language-dependent transcription error rate, instantiated as character error rate (CER) for Chinese and word error rate (WER) for English. 
The group-normalized advantage estimate is given by $\widehat{A}_i=(R_i-\overline{R})/(\sigma_R+\varepsilon)$, where $\overline{R}$ and $\sigma_R$ are the mean and standard deviation of the rewards within the rollout group, respectively. 
Groups with $\sigma_R=0$ are discarded through dynamic sampling. 
DAPO then optimizes the following policy-gradient objective:

\begin{equation}
\mathcal{J}_{\mathrm{DAPO}}(\theta)
=
\mathbb{E}\left[
\frac{
\sum_{i=1}^{G}\sum_{t=1}^{|\tau_i|}
\min\left(
r_{i,t}\widehat{A}_i,\,
\widetilde{r}_{i,t}\widehat{A}_i
\right)
}{
\sum_{i=1}^{G}|\tau_i|
}
\right],
\end{equation}
where $r_{i,t} = \pi_{\theta}(\tau_{i,t}\mid\mathbf{x},\tau_{i,<t})/ \pi_{\theta_{\mathrm{old}}}(\tau_{i,t}\mid\mathbf{x},\tau_{i,<t})$ is the token-level importance ratio, and $\widetilde{r}_{i,t} = \operatorname{clip}(r_{i,t},1-\varepsilon_{\mathrm{low}},1+\varepsilon_{\mathrm{high}})$ is its asymmetrically clipped counterpart. Here, $\varepsilon_{\mathrm{low}}$ and $\varepsilon_{\mathrm{high}}$ denote the clipping margins. Compared with classic RLVR objectives, DAPO uses a token-level objective to balance contributions across transcription lengths and prevent errors in long utterances from being underweighted. Its dynamic sampling strategy further improves efficiency by resampling informative instances.

We construct a fixed 50k-utterance multilingual pool: 20k Mandarin, 10k Mandarin subdialects, 10k Cantonese, and 10k English samples. We group them into three routing categories: Mandarin, Chinese dialects (Mandarin subdialects and Cantonese), and English. Each specialist teacher is trained on all data from its target routing category together with a randomly sampled 20\% subset from each other routing category, yielding three language-specialized teachers, whereas the generalist teacher is trained on the full pool.  
For controlled comparison, DAPO training varies the acoustic front end along two orthogonal dimensions: encoder mode (streaming vs.\ offline) and encoder-adaptor trainability (frozen vs.\ trainable). 

\subsection{Language-Routed Multi-teacher OPD}
\label{sec:language_routed_mopd}

After obtaining the three language-specialized teachers and the generalist teacher, we assess all teachers on held-out validation sets and rank them separately for Mandarin, Chinese dialects (including Mandarin subdialects and Cantonese), and English. Let $\pi^{\mathrm{T}}_{\phi_{\ell(\mathbf{x}),k}}$ denote the $k$-th ranked teacher for routing category $\ell(\mathbf{x})$, where $k=1$ denotes the best-performing teacher. The MOPD student $\pi_\theta$ is initialized from the base model $\pi_{\theta_0}$. 
Following MOPD~\cite{ma2026mopd}, the student generates an on-policy trajectory $\tau\sim\pi_\theta(\cdot\mid\mathbf{x})$. Let $s_t=(\mathbf{x},\tau_{<t})$ denote the decoding state at step $t$. The distillation loss from the $k$-th ranked teacher is defined as
\begin{equation}
\adjustbox{max width=\linewidth}{%
$\displaystyle
\mathcal{L}_{k}(\mathbf{x};\theta)
=
\mathbb{E}_{\tau\sim\pi_\theta}
\left[
\frac{1}{|\tau|}
\sum_{t=1}^{|\tau|}
D_{\mathrm{KL}}\!\left(
\pi_\theta(\cdot\mid s_t)
\,\Vert\,
\pi^{\mathrm{T}}_{\phi_{\ell(\mathbf{x}),k}}
(\cdot\mid s_t)
\right)
\right],
$%
}
\end{equation}
and the final MOPD objective aggregates contributions from the top-$K$ teachers ($K=3$ by default):
\begin{equation}
\label{eq:mopd}
    \mathcal{L}_{\mathrm{MOPD}}(\theta)
    =\mathbb{E}_{\mathbf{x}\sim\mathcal{D}}\!\left[
        \sum_{k=1}^{K}\lambda_k\,\mathcal{L}_{k}(\mathbf{x};\theta)
    \right], \quad\sum_{k=1}^{K}\lambda_k=1.
\end{equation}
The weights $\{\lambda_k\}$ govern the relative influence of each ranked teacher, while setting $\lambda_k=0$ allows the effective number of active teachers to range from three to one. The formulation recovers top-1 teacher distillation (see Fig.~\ref{fig1}(a)) by setting $(\lambda_1,\lambda_2,\lambda_3)=(1,0,0)$, while also supporting weighted multi-teacher aggregation. 
The supervising teacher $\pi^{\mathrm{T}}_{\phi_{\ell(\mathbf{x}),k}}$ is determined according to the evaluation performance of all candidate teachers for the language category $\ell(\mathbf{x})$. This ensures that each student trajectory can be supervised by the empirically strongest teachers for its target language category.


\subsection{Preliminary Analysis on the Effect of Acoustic Prefix}
\label{analysis1}

\begin{figure}[htbp]
    \centering
    \includegraphics[width=1\linewidth]{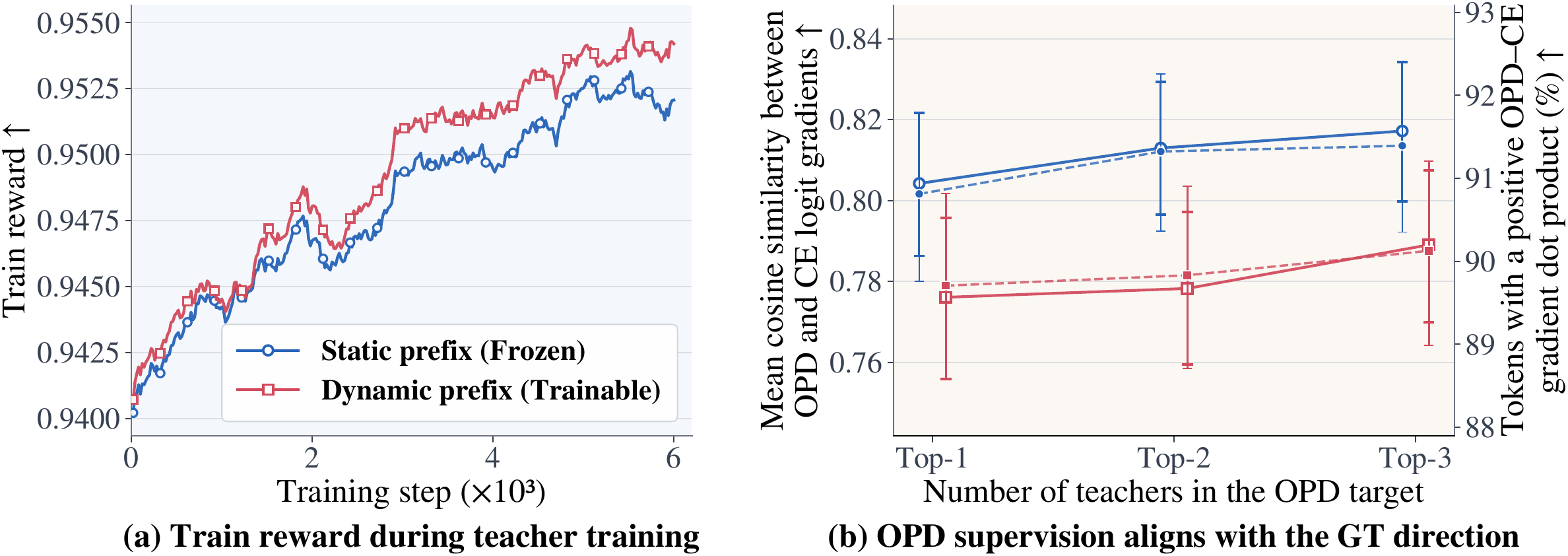}
    \caption{Comparison of static and dynamic acoustic-prefix configurations in terms of (a) accuracy reward trajectories during RL and (b) OPD--CE gradient alignment metrics during distillation. Error bars are estimated from 10k bootstrap resamples.}

    \label{fig2}
    \vspace{-3px}
\end{figure}

When adapting MOPD to ASR, the LLM is conditioned on acoustic embeddings that serve as acoustic prefixes, whose configuration directly influences distillation effectiveness. 
A static prefix configuration ensures prefix consistency between teachers and the student, mitigating the teacher--student conditioning mismatch during distillation. However, this requires all teachers and the student to share a frozen encoder–adaptor throughout RL and OPD, which limits attainable teacher performance. In contrast, a dynamic configuration allows unrestricted joint optimization of all model components, but at the risk of increased conditioning inconsistency.

To illustrate this trade-off, we conduct a preliminary analysis under static and dynamic acoustic-prefix configurations. As shown in Fig.~\ref{fig2}, we compare (a) the accuracy-reward trajectories during RL training and (b) token-level alignment between the OPD gradients and the cross-entropy (CE) gradients derived from ground-truth labels during distillation. The latter is quantified by the mean cosine similarity and the proportion of directionally aligned tokens, defined as those with a positive OPD--CE gradient dot product.  
The results show that the static prefix yields slightly lower teacher performance but better-aligned distillation gradients, whereas the dynamic prefix improves teacher performance at the expense of slightly reduced gradient alignment. This reveals a trade-off between teacher strength and distillation compatibility, further examined in the next section.


\section{Experiments}
\label{sec:experiments}

\begin{table}[!t]
  \centering
  
  \caption{Offline ASR performance measured by CER (\%) for Chinese and WER (\%) for English (lower is better). Model sizes of open-source baselines are shown in the preceding brackets.}
  \label{tab:opd_wer_offline}

  \setlength{\tabcolsep}{2.0pt}
  \renewcommand{\arraystretch}{0.92}

  \resizebox{\columnwidth}{!}{%
  \begin{tabular}{
    @{}
    l
    r@{\,\ensuremath{\vert}\,}l
    c
    r@{\,\ensuremath{\vert}\,}l
    r@{\,\ensuremath{\vert}\,}l
    c
    @{}
  }
    \toprule

    \multirow{2}{*}{\bfseries Method}
    & \multicolumn{2}{c}{\bfseries WeNet}
    & \bfseries KeSpeech
    & \multicolumn{2}{c}{\bfseries WeNet-Yue}
    & \multicolumn{2}{c}{\bfseries LibriSpeech}
    & \multirow{2}{*}{\bfseries Avg.} \\

    & meet. & net
    & test
    & short & long
    & clean & other
    & \\

    \midrule

    \multicolumn{9}{@{}l}{
      \textit{\textbf{Advanced Open-source Models}}
    } \\[0.5pt]

    [0.8B] Fun-ASR-Nano~\cite{an2025fun}
      & 4.68 & 5.22
      & 7.18
      & 7.31 & 10.02
      & 1.63 & 4.35
      & 5.77 \\

    [2.0B] Qwen3-ASR-1.7B~\cite{shi2026qwen3}
      & 4.00 & 4.13
      & 4.98
      & \textbf{5.79} & \textbf{8.00}
      & 1.56 & 3.49
      & \textbf{4.56} \\

    [8B] Step-Audio2-mini~\cite{wu2025step}
      & 4.23 & 4.63
      & \textbf{3.98}
      & 7.78 & 8.44
      & 1.22 & 2.61
      & 4.70 \\

    [30B/A3B] Qwen3-Omni-Inst~\cite{xu2025qwen3}
      & \textbf{3.92} & \textbf{3.85}
      & 5.96
      & 6.97 & 8.60
      & \textbf{1.15} & \textbf{2.38}
      & 4.69 \\

    [8B] MiMo-V2.5-ASR~\cite{coreteam2026mimov25asr}
      & 4.63 & 4.52
      & 7.76
      & 6.53 & 10.52
      & 1.35 & 3.42
      & 5.53 \\

    \midrule

    \multicolumn{9}{@{}l}{
      \textit{\textbf{Our Methods}}
    } \\[0.5pt]

        [2.3B] Base Model (NIM4-ASR)
      & 5.01 & 4.80
      & 4.57
      & 5.22 & 9.48
      & 1.23 & 2.61
      & 4.70 \\

    \addlinespace[1.5pt]

    \multicolumn{9}{@{}l}{
      \confighead{with static acoustic prefix}
    } \\[0.5pt]

    \configmethod{RL (Generalist Teacher)}
      & 4.88 & 4.74
      & 4.40
      & 5.27 & 9.49
      & 1.18 & 2.58
      & 4.65 \\

    \configmethod{RL (Best-Teacher Oracle)}
      & 4.77 & 4.70
      & 4.40
      & 5.19 & 9.30
      & 1.18 & 2.56
      & 4.59 \\

    \configmethod{LS-MOPD with 1 teacher}
      & 4.62 & 4.53
      & \textbf{4.06}
      & 5.19 & \textbf{9.12}
      & 1.15 & 2.57
      & 4.46 \\

    \configmethod{LS-MOPD with 2 teachers}
      & 4.60 & \textbf{4.52}
      & 4.13
      & 5.17 & 9.16
      & 1.15 & 2.53
      & 4.47 \\

    \configmethod{LS-MOPD with 3 teachers}
      & 4.60 & 4.53
      & 4.08
      & \textbf{5.14} & 9.14
      & \textbf{1.12} & 2.56
      & \textbf{4.45} \\

    \addlinespace[1.5pt]

    \multicolumn{9}{@{}l}{
      \confighead{with dynamic acoustic prefix}
    } \\[0.5pt]

    \configmethod{RL (Generalist Teacher)}
      & 4.56 & 4.69
      & 4.49
      & 5.35 & 9.55
      & 1.20 & 2.57
      & 4.63 \\

    \configmethod{RL (Best-Teacher Oracle)}
      & 4.56 & 4.69
      & 4.45
      & 5.25 & 9.19
      & 1.17 & 2.56
      & 4.55 \\

    \configmethod{LS-MOPD with 1 teacher}
      & 4.60 & 4.55
      & 4.14
      & 5.28 & 9.15
      & 1.14 & 2.56
      & 4.49 \\

    \configmethod{LS-MOPD with 2 teachers}
      & \textbf{4.49} & 4.58
      & 4.13
      & 5.23 & 9.16
      & 1.17 & \textbf{2.48}
      & 4.46 \\

    \configmethod{LS-MOPD with 3 teachers}
      & 4.52 & 4.62
      & 4.12
      & 5.17 & 9.27
      & 1.15 & 2.51
      & 4.48 \\

    \bottomrule
  \end{tabular}%
  }

  \parbox{\columnwidth}{
    \footnotesize\raggedright
    \textit{Note: Boldface separately indicates the best result among ``Advanced Open-source Models'' and among ``Our Methods'' for each benchmark.}
  }
\end{table}

\subsection{Experimental Setup}

Our base student model is built upon an in-house backbone pretrained on approximately 560k hours of labeled speech, comprising a 0.6B FireRed Conformer~\cite{xu2026fireredasr2s}, a $4\times$ downsampling linear adaptor, and a Qwen3-1.7B LLM decoder. 
For RL and MOPD, we construct a 50k-utterance multilingual training pool by sampling from representative datasets: 20k Mandarin utterances from WenetSpeech~\cite{zhang2022wenetspeech}, 10k Mandarin subdialect utterances from KeSpeech~\cite{tang2021kespeech}, 10k Cantonese utterances from WenetSpeech-Yue~\cite{li2026wenetspeech}, and 10k English utterances from LibriSpeech~\cite{panayotov2015librispeech}. The generalist teacher and MOPD stages use the full multilingual pool, whereas language-specialized teachers are trained on their respective language-skewed subsets constructed with sampling ratios specified in Section~\ref{sec:domain_specialized_rl}.

For evaluation, we employ the official test sets of the aforementioned four datasets. Across all evaluations, a unified text normalization pipeline based on WeTextProcessing\footnote{https://github.com/wenet-e2e/WeTextProcessing} is applied to mitigate effects of surface-form variations. 
All evaluations use deterministic decoding strategies. In the streaming setting, speech is encoded in 640 ms chunks, with four preceding chunks retained as left context, and greedy decoding (beam size = 1) is employed. In the offline setting, the encoder operates on the complete utterance, and beam search decoding is applied with a beam size of 3.


\subsection{Training Details}

For RL training, we train for at most 6k optimization steps with a maximum learning rate of $2\times10^{-6}$ linearly decayed to zero, and a sampling temperature that is cosine-annealed from 1.0 to 0.7. For MOPD, we train for 20k steps with the learning rate cosine-annealed from \(4\times10^{-6}\) to \(2\times10^{-6}\) and a fixed sampling temperature of 1.0. All RL and MOPD experiments are conducted on 8 NVIDIA A100 GPUs with BF16 precision and DeepSpeed ZeRO-2~\cite{rajbhandari2020zero}.

\subsection{Main Results}

\begin{table}[!t]
  \centering

  \caption{Streaming ASR performance measured by CER (\%) for Chinese and WER (\%) for English (lower is better).}
  \label{tab:opd_wer_streaming}

  \setlength{\tabcolsep}{2.0pt}
  \renewcommand{\arraystretch}{0.92}

  \resizebox{\columnwidth}{!}{%
  \begin{tabular}{
    @{}
    l
    r@{\,\ensuremath{\vert}\,}l
    c
    r@{\,\ensuremath{\vert}\,}l
    r@{\,\ensuremath{\vert}\,}l
    c
    @{}
  }
    \toprule

    \multirow{2}{*}{\bfseries Method}
    & \multicolumn{2}{c}{\bfseries WeNet}
    & \bfseries KeSpeech
    & \multicolumn{2}{c}{\bfseries WeNet-Yue}
    & \multicolumn{2}{c}{\bfseries LibriSpeech}
    & \multirow{2}{*}{\bfseries Avg.} \\

    & meet. & net
    & test
    & short & long
    & clean & other
    & \\

    \midrule

    Base Model (NIM4-ASR)
      & 6.19 & 5.26
      & 5.69
      & 6.06 & 12.33
      & 1.34 & 3.32
      & 5.74 \\

    \addlinespace[1.5pt]

    \multicolumn{9}{@{}l}{
      \confighead{with static acoustic prefix}
    } \\[0.5pt]

    \configmethod{RL (Generalist Teacher)}
      & 5.92 & 5.19
      & 5.54
      & 5.68 & 11.10
      & 1.29 & 3.27
      & 5.43 \\

    \configmethod{RL (Best-Teacher Oracle)}
      & 5.84 & 5.18
      & 5.54
      & \textbf{5.50} & 11.10
      & 1.25 & 3.20
      & 5.37 \\

    \configmethod{LS-MOPD with 1 teacher}
      & 5.76 & 5.00
      & 5.12
      & 5.61 & 10.83
      & 1.25 & 3.21
      & 5.25 \\

    \configmethod{LS-MOPD with 2 teachers}
      & 5.67 & 4.99
      & 5.17
      & 5.59 & 10.67
      & 1.24 & 3.15
      & 5.21 \\

    \configmethod{LS-MOPD with 3 teachers}
      & 5.69 & \textbf{4.96}
      & \textbf{5.09}
      & 5.52 & \textbf{10.61}
      & \textbf{1.23} & 3.19
      & \textbf{5.18} \\

    \addlinespace[1.5pt]

    \multicolumn{9}{@{}l}{
      \confighead{with dynamic acoustic prefix}
    } \\[0.5pt]

    \configmethod{RL (Generalist Teacher)}
      & 5.90 & 5.26
      & 5.52
      & 5.60 & 11.10
      & 1.35 & 3.22
      & 5.42 \\

    \configmethod{RL (Best-Teacher Oracle)}
      & 5.72 & 5.26
      & 5.50
      & 5.58 & 10.96
      & 1.31 & 3.16
      & 5.36 \\

    \configmethod{LS-MOPD with 1 teacher}
      & \textbf{5.65} & 5.17
      & 5.30
      & 5.58 & 11.19
      & 1.29 & 3.17
      & 5.34 \\

    \configmethod{LS-MOPD with 2 teachers}
      & 5.71 & 5.16
      & 5.24
      & 5.99 & 11.02
      & 1.27 & \textbf{3.12}
      & 5.36 \\

    \configmethod{LS-MOPD with 3 teachers}
      & 5.78 & 5.26
      & 5.22
      & 5.51 & 10.90
      & 1.28 & 3.15
      & 5.30 \\

    \bottomrule
  \end{tabular}%
  }
\end{table}

Tables~\ref{tab:opd_wer_offline} and~\ref{tab:opd_wer_streaming} report the performance of our proposed methods on benchmarks, alongside leading open-source models~\cite{an2025fun,shi2026qwen3,wu2025step,xu2025qwen3,coreteam2026mimov25asr} as references. 
Within each acoustic-prefix configuration, the generalist teacher serves as the RL baseline, whereas the best result attained by any candidate teacher on each benchmark is reported as ``RL (Best-Teacher Oracle)''.
For LS-MOPD, the number of supervising teachers is determined by the loss weights in Eq.~\eqref{eq:mopd}: the one-, two-, and three-teacher variants use $\mathcal{L}_{\mathrm{MOPD}}=\mathcal{L}_1$, $\mathcal{L}_{\mathrm{MOPD}}=0.6\mathcal{L}_1+0.4\mathcal{L}_2$, and $\mathcal{L}_{\mathrm{MOPD}}=0.6\mathcal{L}_1+0.2\mathcal{L}_2+0.2\mathcal{L}_3$, respectively. We analyze the results as follows:

\textbf{(1) Overall Performance.} Using only a 50k-utterance post-training pool, LS-MOPD substantially improves upon the base model and consistently outperforms the RL baseline. With only 2.3B parameters, LS-MOPD achieves a minimum average error rate of 4.45\%, outperforming all open-source baselines, including models trained on million-hour-scale speech corpora~\cite{an2025fun,shi2026qwen3} and those built upon larger LLM backbones~\cite{wu2025step,xu2025qwen3,coreteam2026mimov25asr}. Notably, LS-MOPD surpasses the empirical performance envelope defined by the best-teacher oracle on almost all benchmarks, and achieves the best results among all compared leading systems on WeNetSpeech-Yue-short and LibriSpeech test-clean. These results demonstrate that the distilled student can outperform all teachers, achieving highly competitive multilingual recognition accuracy together with measurable emergent gains. 
As shown by streaming results in Table~\ref{tab:opd_wer_streaming}, LS-MOPD also delivers substantial gains in the streaming setting, with relative improvements exceeding those in the offline configuration.

\textbf{(2) Static vs. dynamic acoustic prefix.} Before distillation, dynamic-prefix RL teachers achieve slightly better average performance than their static-prefix counterparts, owing to joint encoder--adaptor optimization that yields more refined acoustic representations. After MOPD, however, static-prefix variants attain the best average results in both the offline and streaming settings. We attribute this reversal to teacher--student acoustic-prefix mismatch under the dynamic configuration: when teacher supervision depends on acoustic cues absent from the student's representation, distillation may shift from learnable preference transfer toward fitting partially unreachable targets. 

\textbf{(3) Top-$K$ teacher weighted aggregation.} Since cross-lingual heterogeneity is generally less pronounced than cross-task heterogeneity, we extend the default top-1 routing of MOPD to weighted top-$K$ teacher aggregation. This yields modest gains, particularly in the streaming setting, suggesting a beneficial regularization effect that becomes more pronounced under higher decoding uncertainty. Notably, we find that incorporating the second- and third-ranked teachers can lead to improvements despite their substantially lower standalone performance than the best teacher, suggesting that teacher diversity can provide complementary supervision and alleviate token-level overconfidence.

\subsection{Sources of Improvement and Emergent Gains}

\begin{figure}[!h]
    \centering
    \includegraphics[width=1.0\linewidth]{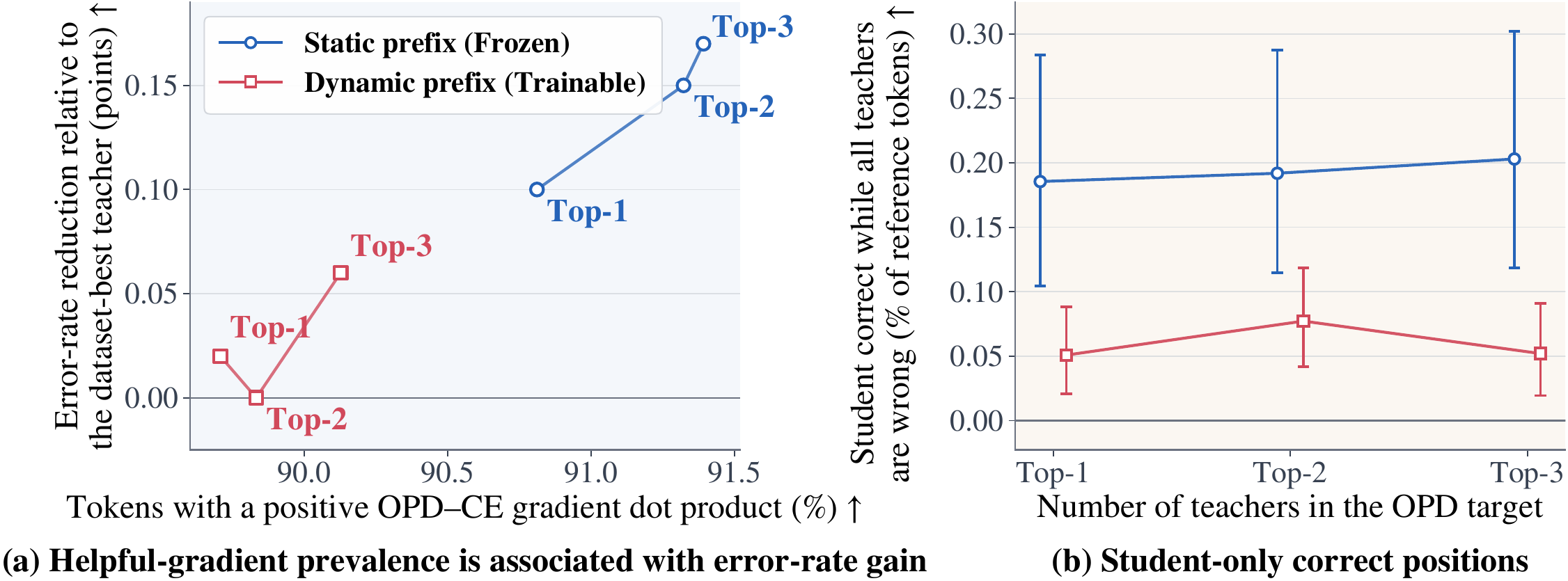}
    \caption{(a) Relationship between OPD--CE gradient alignment and the performance gains achieved by LS-MOPD. (b) Emergent gains: the proportion of tokens incorrectly predicted by all teachers but correctly predicted by the student.}
    \label{fig3}
    \vspace{-3px}
\end{figure}

Beyond benchmark results, we conduct diagnostic analyses to validate LS-MOPD and offer insight into its emergent behavior. As shown in Fig.~\ref{fig3}(a), we examine the relationship between benchmark gains of different LS-MOPD variants and the proportion of tokens with aligned OPD--CE gradients, as defined in Section~\ref{analysis1}. 
The results show a positive association: static-prefix LS-MOPD variants exhibit more consistently aligned gradients and achieve larger performance gains. By contrast, although dynamic-prefix teachers show stronger standalone performance, their supervision induces gradients that are less aligned with the CE gradients, leading to smaller distillation gains. 
Moreover, multi-teacher supervision generally yields better gradient alignment than single-teacher supervision. This finding provides a preliminary mechanistic explanation for why multi-teacher distillation can outperform its single-teacher counterpart: lower-ranked teachers can still contribute complementary supervision that induces more favorable optimization directions.

Fig.~\ref{fig3}(b) further reveals the emergent benefit of LS-MOPD: a non-negligible proportion of tokens are predicted incorrectly by all teachers, yet are correctly recognized by the student. Together with the consistent improvements of LS-MOPD over the best-teacher oracle in Tables~\ref{tab:opd_wer_offline} and~\ref{tab:opd_wer_streaming}, these results highlight the potential of LS-MOPD for multilingual ASR, demonstrating that coordinated supervision from multiple teachers can enable the student to generalize beyond the performance envelope of any individual teacher.



\section{Conclusion}
\label{sec5}

This work proposes a multi-teacher on-policy distillation framework for multilingual LLM-based ASR. We systematically investigate the impact of static and dynamic acoustic-prefix configurations and integrate language-specialized teachers through language-routed, weighted supervision. Experimental results across both offline and streaming scenarios demonstrate that LS-MOPD substantially outperforms RL baselines and has the potential to generalize beyond the empirical upper envelope of trained RL teachers. We further analyze the effectiveness of LS-MOPD from the perspectives of gradient alignment and multi-teacher complementarity, and derive practical insights into acoustic-prefix selection and multi-teacher aggregation.

Our work also has several limitations. First, the evaluation covers only a single model backbone and a limited set of languages, warranting broader validation across architectures and languages. Second, the dynamic acoustic-prefix configuration remains preliminary and does not fully exploit its potential. Future work will explore teacher--student supervision mechanisms that retain the optimization flexibility while reducing conditioning mismatch.

\clearpage
\bibliographystyle{IEEEbib}
\bibliography{refs}

\end{document}